\documentclass[runningheads]{llncs}
\usepackage{amsfonts}
\usepackage{graphicx}

\usepackage{wrapfig} 
\usepackage{amsmath}
\usepackage{float}   
\usepackage{array}
\usepackage{caption}
\usepackage{booktabs}
\usepackage{xcolor}
\usepackage{adjustbox}
\usepackage{rotating} 

\begin{document}
\title{FoCLIP: A Feature-Space Misalignment Framework for CLIP-Based Image Manipulation and Detection}
\titlerunning{FoCLIP}
%
%
\newcommand{\equal}{\textsuperscript{\dag}}   
\newcommand{\corr}{\textsuperscript{*}}       

\author{
  Yulin Chen\inst{1}
  \and
  Zeyuan Wang\inst{1}
  \and
  Tianyuan Yu\inst{1}\corr
  \and
  Yingmei Wei\inst{1}
  \and
  Liang Bai\inst{1}
}

\authorrunning{Y. Chen et al.}

\institute{
  Laboratory for Big Data and Decision, National University of Defense Technology, Changsha 410073, China
}

\maketitle              
\begin{abstract}
The well-aligned attribute of CLIP-based models enables its effective application like CLIPscore as a widely adopted image quality assessment metric.
However, such a CLIP-based metric is vulnerable for its delicate multimodal alignment. 
In this work, we propose \textbf{FoCLIP}, a feature-space misalignment framework for fooling CLIP-based image quality metric. 
Based on the stochastic gradient descent technique, FoCLIP integrates three key components to construct fooling examples: feature alignment as the core module to reduce image-text modality gaps, the score distribution balance module and pixel-guard regularization, which collectively optimize multimodal output equilibrium between CLIPscore performance and image quality.
Such a design can be engineered to maximize the CLIPscore predictions across diverse input prompts, despite exhibiting either visual unrecognizability or semantic incongruence with the corresponding adversarial prompts from human perceptual perspectives.
Experiments on ten artistic masterpiece prompts and ImageNet subsets demonstrate that optimized images can achieve significant improvement in CLIPscore while preserving high visual fidelity. 
In addition, we found that grayscale conversion induces significant feature degradation in fooling images, exhibiting noticeable CLIPscore reduction while preserving statistical consistency with original images. 
Inspired by this phenomenon, we propose a color channel sensitivity-driven tampering detection mechanism that achieves 91\% accuracy on standard benchmarks.
In conclusion, this work establishes a practical pathway for feature misalignment in CLIP-based multimodal systems and the corresponding defense method.
\end{abstract}
\begingroup
\renewcommand\thefootnote{}
\footnotetext{\corr\ Corresponding author: Tianyuan Yu (\email{tianyuan.yu@nudt.edu.cn}).}
\endgroup
\section{Introduction}
With the rapid development of artificial intelligence technology, multimodal learning, a bridge that connects different types of data (such as images and text), has become an important research direction in the field of artificial intelligence\cite{lin2024preflmr,yu2024recent}. The CLIP (Contrastive Language-Image Pre-training) model\cite{radford2021learning},
\begin{wrapfigure}{r}{0.5\textwidth}
    \centering
    \vspace{-0.7cm}
    \includegraphics[trim=330 130 270 90, clip,width=.6\textwidth, page=1]{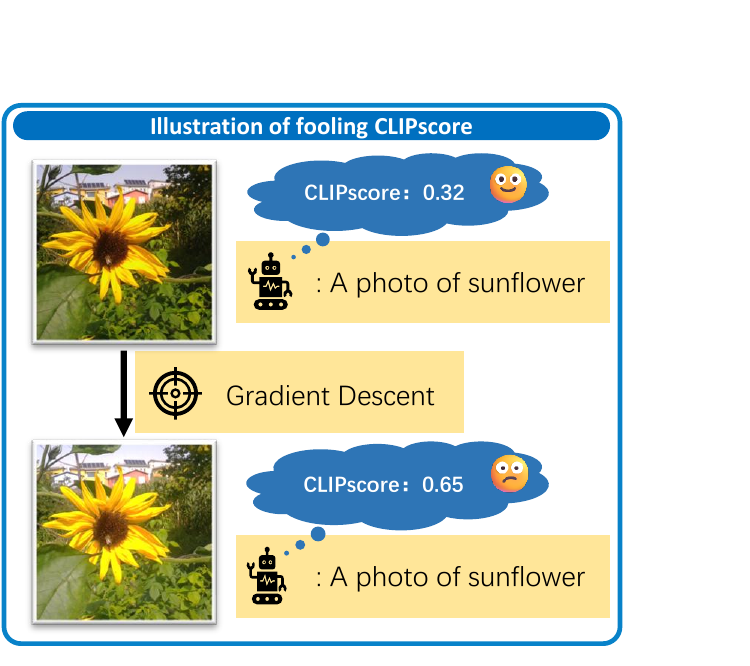} 
    \caption{Illustration of fooling CLIPscore. As shown, 0.32 is the correct score, but through our FoCLIP method, despite this being visually inconsistent, the CLIPscore is unexpectedly high.}
    \vspace{-0.7cm}
    \label{fig:Fool}
\end{wrapfigure}
as a classic language-image contrast model, has received widespread attention due to its excellent performance in multimodal tasks\cite{xu2025toward,xie2025science}. The CLIP model realizes cross-modal information fusion by learning the association between images and text, providing strong technical support for applications such as image retrieval and image-text matching\cite{ramesh2021zero,yang2022chinese,jina2024clip}.\\
\indent However, the widespread deployment of the CLIP model in practical applications made security issues gradually emerged. Studies have shown that the CLIP model is vulnerable to malicious attacks\cite{bai2024badclip,chen2023resonance,han2025detect,li2024bdetclip}. Attackers can use carefully designed tampering methods to disrupt the correct matching relationship between images and text, thereby misleading the model. Some studies\cite{liang2022mind,FreibergeCLIPMasterPrints} show that this vulnerability of the CLIP model is very likely to come from a modality gap between text and image embeddings. This vulnerability not only limits the application of the CLIP model in scenarios with high security requirements\cite{chen2023resonance,han2025detect,li2024bdetclip}, but also poses a challenge to the authenticity and integrity of digital content.\\
\indent Based on the research of Matthias Freiberger et al.\cite{FreibergeCLIPMasterPrints}, this paper proposes a feature-space misalignment framework FoCLIP. FoCLIP realizes the directional enhancement of specific semantic concepts by jointly optimizing the feature distribution of the image in the CLIP multimodal embedding space while maintaining the visual quality of the image. In addition, this study also explores the sensitivity phenomenon of the CLIP feature-optimized image in the color space conversion and proposes a detection mechanism based on grayscale sensitivity to detect whether the image has been tampered with. Experimental results show that FoCLIP can not only effectively improve the CLIPscore of the image to fool the CLIP model, but also achieve high-accuracy tampering detection on the standard test set.
The main contributions of this paper include:\\
\indent\textbf{1. Multi-objective Joint Optimization}: A tripartite optimization framework, FoCLIP, was developed that integrates Feature Alignment loss, Distribution Balance loss, and Pixel-Guard Regularization loss. This framework achieves alignment with the target prompts (42.7\% average improvement in CLIPscore on artistic prompts) while preserving image quality.\\
\indent\textbf{2. Robust Generalization}: Comprehensive experiments demonstrated FoCLIP's stable generalization across 25-100 class scales on ImageNet, showing a 27.3\% average CLIPscore improvement.\\
\indent\textbf{3. Color Channel Sensitivity Discovery}: We discovered the vulnerability in which the scores of images generated by the deception method for CLIPscore significantly decreased after grayscale conversion through experiments. Based on this finding, a grayscale sensitivity-based detection mechanism was proposed, which achieves 91\% accuracy in tampering detection in the ImageNet validation set (Section 4.3).

\section{Related work}
\subsection{Adversarial Attacks in Multimodal Models}
With the rapid advancement of artificial intelligence technologies, in recent years there have been growing concerns about the security of multimodal models\cite{yang2024safebench,zhu2025multitrust,jiang2025multimodal,chen2025security}. The CLIP model proposed by Radford et al.\cite{radford2021learning}, achieves cross-modal feature alignment through contrastive learning, but its open feature space introduces security risks\cite{qiu2022multimodal,noever2021reading,daras2023discovering,goh2021multimodal}. Studies demonstrate that attackers can compromise cross-modal consistency via gradient optimization. Dong et al.\cite{dong2018boosting} employed Projected Gradient Descent (PGD) to generate adversarial examples that successfully induce high confidence mismatches in CLIP for manipulated images. Qin et al.\cite{qin2018black} developed a black-box optimization framework that allows malicious cross-modal association control through API access only. These findings reveal the vulnerability of multimodal systems to adversarial attacks, although existing defenses predominantly address single-modal scenarios, failing to counter the coupled characteristics of cross-modal attacks.
\subsection{Image Tampering Detection Techniques}
Traditional digital forensics is based on the analysis of characteristics of the physical layer. Farid's EXIF metadata verification\cite{farid2009exposing} detects file header anomalies, but remains ineffective against semantic-level content tampering. In the deep learning era, Zhang et al.\cite{zhang2018learning} utilized CNNs to extract frequency domain features, allowing the detection of JPEG compression and copy-move forgeries. For adversarial examples, Xu et al.\cite{xu2021detecting} proposed a feature map discrepancy analysis, which compares activation patterns between clean and adversarial samples. However, these methods show limited efficacy against cross-modal attacks, which struggle to capture feature shifts in semantic space.
\subsection{Multimodal Feature Alignment Methods}
Cross-modal alignment constitutes the core challenge in multimodal learning. The contrastive loss proposed by Chen et al.\cite{chen2020simple} maximizes the positive pair similarity, but suffers from coarse semantic granularity. Wang et al.\cite{wang2019cross} introduced hierarchical attention mechanisms for fine-grained image-text matching. Recent advances explore adversarial alignment optimization. Jia et al.\cite{jia2021scaling} applied orthogonal constraints in feature space to prevent modal dominance, while Liang et al.\cite{liang2022multi} developed dynamic weight allocation to balance multi-objective conflicts. Despite improved alignment performance, these approaches lack robustness guarantees in adversarial environments.

\section{Methodology}
\begin{figure}[htbp]
  \centering
  \vspace{-0.7cm}
  \includegraphics[trim=0 40 0 20, clip,width=1\textwidth, page=1]{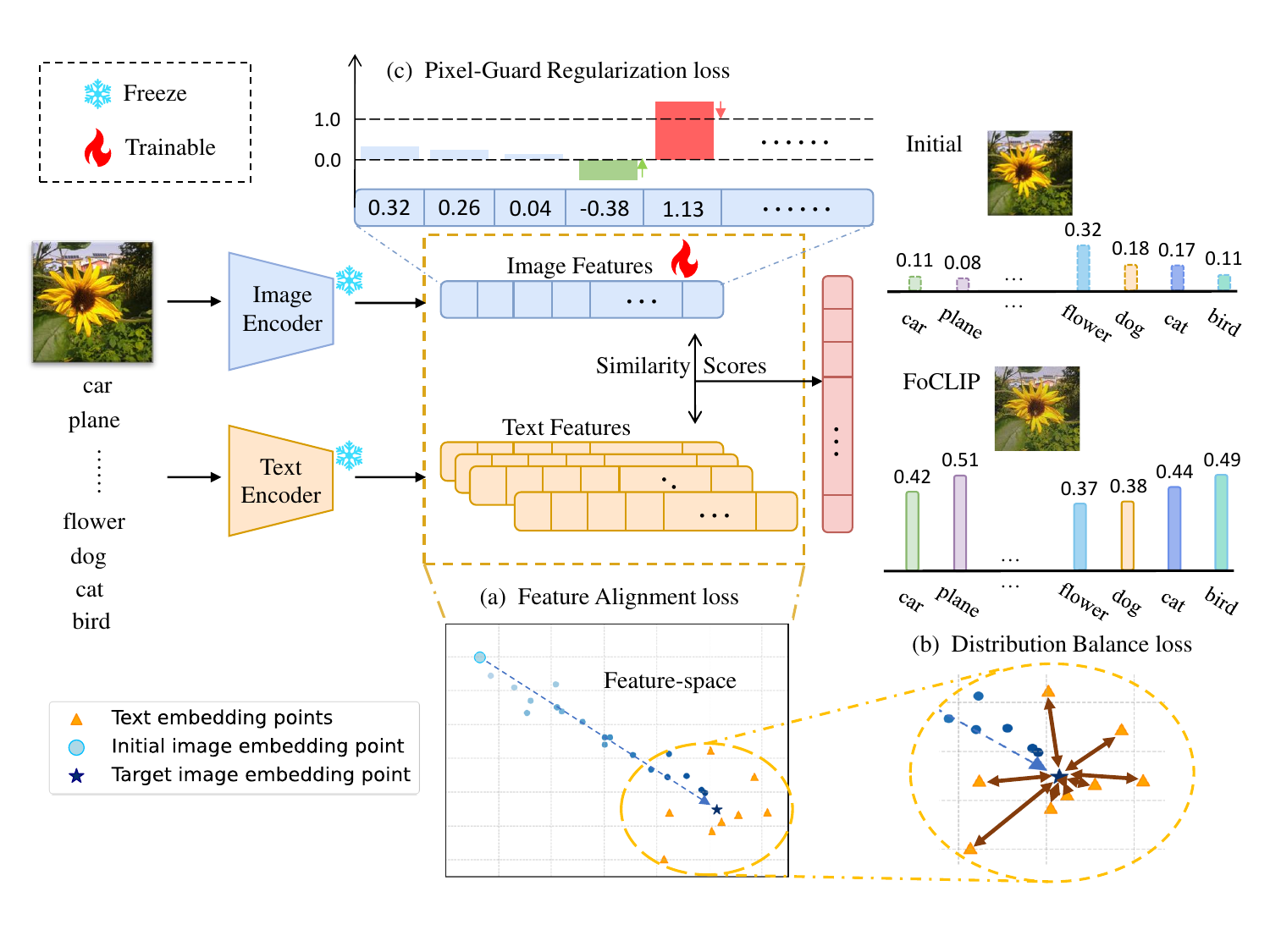} 
  \caption{The framework of FoCLIP, a tripartite optimization approach for adversarial CLIPscore manipulation. Built upon stochastic gradient descent (SGD) updates to the image feature vector $ \mathbf{g}(x) $, this framework iteratively adjusts pixel values to bridge the modality gap between visual and textual embeddings. The architecture decomposes the adversarial process into three synergistic components: (a) Feature Alignment Loss minimizes the cosine distance between image features and target text prompts to enhance semantic alignment in CLIP's embedding space. (b) Distribution Balance Loss ensures balanced similarity scores across multiple prompts by penalizing variance, avoiding overfitting to specific concepts. (c) Pixel-Guard Regularization Loss constrains pixel values within a predefined range $[bound_{lower}, bound_{upper}]$ via ReLU limitations, preserving visual fidelity during optimization.}
  \label{fig:mainfig}
  \vspace{-0.4cm}
\end{figure}

\noindent Fig.\ref{fig:mainfig} provides an overview of the FoCLIP framework, which leverages feature-space misalignment to systematically optimize CLIPscore while maintaining image quality. By decomposing the adversarial process into three synergistic modules: feature alignment, distribution balance, and pixel-level regularization. we achieve a targeted enhancement of multimodal alignment strength without compromising image quality. This architecture explicitly addresses the modality gap inherent in CLIP-based metrics, enabling both effective fooling of CLIP models and subsequent tampering detection through color-channel sensitivity analysis (see Section 3.2). The following subsections detail the mathematical formulation of each component, forming a tripartite equilibrium framework for adversarial CLIPscore manipulation and a grayscale sensitivity detection mechanism.

\subsection{Feature-Space Misalignment Framework Based on Pre-trained Models}
In general, the pre-trained CLIP model can indicate how well a prompt or image caption $c$ aligns with the given image $x$. For a given prompt-image pair $(c,\mathbf{x})$, $\mathcal{C}_\theta$ extracts a pair of corresponding feature embeddings $(f(c), g(\mathbf{x}))$ and computes their cosine similarity: 
\begin{equation}
s(\mathbf{x}, c)=\mathcal{C}_\theta(\mathbf{x}, c)=\frac{\mathbf{g}(\mathbf{x})^{\top}}{\|\mathbf{g}(\mathbf{x})\|} \cdot \frac{\mathbf{f}(c)}{\|\mathbf{f}(c)\|}
\label{equ:1}
\end{equation}
where  the score $s(\mathbf{x}, c)$ ranges between [0, 1], where values closer to 1 indicate higher image-text alignment, while those nearer to 0 suggest poorer alignment. In practice, it has been found that $s(x, c) = 1$ is hardly achieved, and even for well-fitting text-image pairs $s(x, c) \leq 0.3$, indicating that the underlying misalignment between image and text embeddings as the modality gap of the CLIP-based metric \cite{liang2022mind}.  In this work, our aim is to exploit this modality gap to explore fooling master images by means of stochastic gradient descent. So that we can find the embedding $\mathbf{g}(\mathbf{x_{Fo}})$ of a master image $\mathbf{x_{Fo}}$ for a number of matching prompt-image pairs $\left(c_1, \mathbf{x}_1\right),\left(c_2, \mathbf{x}_2\right), \ldots\left(c_n, \mathbf{x}_n\right)$ such that: 
\begin{equation}
\frac{\mathbf{g}\left(\mathbf{x}_{\text {Fo}}\right)^{\top}}{\left\|\mathbf{g}\left(\mathbf{x}_{\text {Fo}}\right)\right\|} \cdot \frac{\mathbf{f}\left(c_k\right)}{\left\|\mathbf{f}\left(c_k\right)\right\|}>\frac{\mathbf{g}\left(\mathbf{x}_k\right)^{\top}}{\left\|\mathbf{g}\left(\mathbf{x}_k\right)\right\|} \cdot \frac{\mathbf{f}\left(c_k\right)}{\left\|\mathbf{f}\left(c_k\right)\right\|} \quad \text { for } \quad k \in[1, n]
\end{equation}
The most straightforward approach to achieve this target is to maximize equation (\ref{equ:1}) by means of stochastic gradient descent (SGD). This gradient $\nabla_{\mathbf{x}}\left(-s\left(\mathbf{x}, c\right)\right)$  is direct but vulnerable, which can cause the target embedding to collapse. So, our loss function $\mathcal{L}$ is enriched into three parts: feature alignment loss $\mathcal{L}_{align}$, distribution balance loss $\mathcal{L}_{var}$, and the pixel-guard regularization loss $\mathcal{L}_{pixel}$. Among them, the feature alignment loss is the core, which is used to learn to reduce the modality gap between images and texts. The distribution balance and pixel-guard losses regularize optimization to balance CLIPscore improvement and visual quality. The total loss function of our method is as follows:
\begin{equation}
\label{equ: loss_all}
\mathcal{L} =  \mathcal{L}_{align} + \alpha \cdot\mathcal{L}_{var} + \beta\ \cdot \mathcal{L}_{pixel}
\end{equation}
where \(\alpha\) and $\beta$ is the weighting coefficient to balance the influence of different loss component.
\subsubsection{Feature Alignment Loss}
The feature alignment component minimizes the cosine distance between the image features and the set of target text prompts, thereby enhancing the features related to specific semantic concepts in the image, such that the image representation in the CLIP model becomes closer to the representations of target texts, by computing the cosine similarity between the image feature vector $g(x)$ and each text prompt feature vector $f(c_i)$, then averaging the similarities across all prompts and taking the negative value to derive the loss: 
\begin{equation}
\mathcal{L}_{align} = -\frac{1}{N}\sum_{i=1}^Ns(\mathbf{x}, c_i) = \frac{1}{N}\sum_{i=1}^N\left(- \frac{g(x)^\top f(c_i)}{\|g(x)\|\|f(c_i)\|}\right) 
\end{equation}
The feature alignment component function is used to ensure that the optimized image is aligned with the feature distribution of the target semantic concept in the multimodal embedding space of CLIP.  \(N\) is the number of targeted text prompts. \(g(x)\) is the image feature vector extracted by the CLIP model, representing the embedding of the input image x in the CLIP image encoder. \(f(c_i)\) is the vector of text characteristics extracted by the CLIP model, representing the embedding of the \(i^{th}\) text prompt \(c_i\) in the CLIP text encoder. \(\frac{g(x)^\top f(c_i)}{\|g(x)\|\|f(c_i)\|}\) is the cosine similarity between the image feature vector and the text feature vector, and a lower value indicates greater similarity of features.
\subsubsection{Distribution Balance Loss}
\indent The distribution balance loss is used to balance the feature distribution of the optimized image to make it closer to the distribution of the target semantic concept, by calculating the variance of the cosine similarity $s(x,c_i)$ between the image and all text prompts, which can be denoted as: 
\begin{equation}
\mathcal{L}_{var} =  \text{Var}(\{s(x,c_i)\}_{i=1}^N) 
\end{equation} 
This term is used to prevent the optimization process from overly favoring certain specific prompts. \(s(x,c_i)\) is the cosine similarity between image x and \(i^{th}\) text prompt \(c_i\). The \(\alpha\) in equation (\ref{equ: loss_all}) is the regularization coefficient, used to control the weight of the distribution balance loss. 
\subsubsection{Pixel-Guard Regularization Loss}
\indent The pixel-guard regularization component is represented by $\mathcal{L}_{pixel}$, and this part of the loss function is used to ensure that the optimized image pixel values are within a reasonable range. This involves the constraint of pixel values to prevent unnatural image changes during the optimization process, by calculating the average value of the ReLU function of all pixel values to obtain the loss. The ReLU function sets negative values to a preset value and keeps positive values unchanged, thereby penalizing pixel values that exceed the range:
\begin{equation}
\mathcal{L}_{pixel} = \mathbb{E}\left[\text{ReLU}(x-bound_{upper}) + \text{ReLU}(bound_{lower}-x)\right] 
\end{equation}
Here, $[bound_{lower} , bound_{upper}]$ represents a range. Applying the Rectified Linear Unit $ReLU(\cdot)$ function and averaging all pixel values ensure that the optimized image pixel values remain within the expected range.

\subsubsection{Theoretical Gradient Analysis and Misalignment Mechanism}
Let $s(x,c_i)=\langle \hat g(x),\hat f(c_i)\rangle$, define $\bar f = \frac{1}{N}\sum_{i=1}^{N} \hat f(c_i)$ and $\hat g(x)=g(x)/\|g(x)\|$. Then the feature alignment loss can be rewritten as $\mathcal{L}_{\text{align}}=-\langle \hat g(x),\bar f\rangle$, yielding the gradient on the unit sphere $\nabla_{\hat g} \mathcal{L}_{\text{align}} = -\bar f + \langle \hat g,\bar f\rangle\hat g$. This update lies in the tangent space (orthogonal to $\hat g$), nudging image features along the semantic arc towards $\bar f$ while introducing tangential drift that moves $g(x)$ off the natural-image manifold with minimal pixel-space deformation.
For distribution balance $\mathcal{L}_{\text{var}}=\text{Var}({s_i}) = \frac{1}{N}\sum_i s_i^2 - (\frac{1}{N}\sum_i s_i)^2$, its gradient can be written as $\nabla_g \mathcal{L}_{\text{var}} = \frac{2}{N}\sum_i (s_i-\bar s)\nabla_g s_i$, penalizing outlier prompts and, on the hypersphere, acting like an isotropic Dirichlet-style prior that pushes $\hat g$ towards the angular barycenter of prompts. This amplifies cross-prompt projection differences and further reinforces misalignment while still increasing the overall CLIPscore.
For pixel-guard $\mathcal{L}_{\text{pixel}}=\mathbb{E}[\mathrm{ReLU}(x-b_u)+\mathrm{ReLU}(b_l-x)]$, the gradient is nonzero only for out-of-bound pixels, forming a sparse mask $\nabla_x \mathcal{L}_{\text{pixel}} = \mathbf{1}_{x>b_u}-\mathbf{1}_{x<b_l}$. It suppresses large pixel drifts and concentrates updates in high-frequency color/texture subspaces that CLIP is particularly sensitive to. Superimposed with FA/DB’s semantic gradients, this creates tension between semantic alignment and visual fidelity, yielding sizable embedding shifts under small, unobtrusive pixel edits.\
In summary, the three losses drive $g(x)$ away from the native manifold through three approximately orthogonal pathways: semantic pointing $\rightarrow$ subspace tangent drift; multi-prompt balancing $\rightarrow$ angle-center drift; pixel box constraints $\rightarrow$ color high-frequency drift. This explains why FoCLIP can substantially improve CLIPscore while preserving visual quality, and aligns with reports that CLIP’s training may introduce cross-modal bias amplification.

\subsection{Grayscale sensitivity detection}
During the experiments, we observed that although the optimized images achieve high CLIPscore, converting them into grayscale causes a significant drop in similarity while preserving low-level statistics. Inspired by this phenomenon, we propose a double-threshold detection mechanism: when the original image maintains a high similarity while its grayscale counterpart drops below a threshold, the sample is flagged as tampered.\\
\indent Absolute threshold:
\begin{equation}
D(x) > \tau_1
\end{equation}
\indent Relative threshold:
\begin{equation}
\frac{D(x)}{s(x)} > \tau_2
\end{equation}
\indent We quantify color-channel dependence via the grayscale sensitivity difference:
\begin{equation}
D(x)=\frac{1}{N}\sum^{N}_{i=1}\left|s(x,c_i)-s(\mathrm{Gray}(x),c_i)\right|
\end{equation}

Among them, N represents the number of target text prompts, $s(x,c_i)$ represents the cosine similarity between image x and the $i^{th}$ text prompt $c_i$, Gray(x) refers to converting image x into a grayscale image, $s(\text{Gray}(x),c_i)$ refers to the cosine similarity between the grayscale image Gray(x) and the $i^{th}$ text prompt $c_i$, $\left|s(x,c_i) - s(\text{Gray}(x),c_i)\right|$is used to calculate the absolute value of the difference in cosine similarity between the original image and the grayscale image with each text prompt, and $\frac{1}{N}\sum_{i=1}^N$ means to take the average of the similarity differences for all target text prompts.

\paragraph{Robustness and Evasiveness Analysis.} The dual-threshold rule flags tampering when $D(x)>\tau_1$ and $\frac{D(x)}{s(x)}>\tau_2$ while the sample maintains a high target similarity $s(x)>\theta$. An adaptive attacker must therefore simultaneously: (i) keep $D(x)$ small, (ii) keep $\frac{D(x)}{s(x)}$ small, and (iii) keep $s(x)$ large -- three mutually conflicting constraints. Suppressing $D(x)$ typically requires abandoning the color-channel directions that most increase $s(x)$, whereas increasing $s(x)$ tends to enlarge $D(x)$, creating a trade-off that raises the optimization cost of evasive attacks. Our detector relies only on native CLIP features and introduces no trainable parameters, and it is compatible with additional lightweight consistency checks (like JPEG consistency and color-jitter/channel-shuffle rescoring).

\section{Experiments}
\subsection{Experimental Setup}
We followed Matthias Freiberger’s experimental setup \cite{FreibergeCLIPMasterPrints}, testing our method in CLIP ($ViT-L/14@336px$) for famous artwork and ImageNet. For artworks, we trained a fooling master image to maximize CLIPscore across 10 prompts (titles/authors of famous artworks), using the original sunflower image and the SGD optimizer. Initial optimization ran 1,000 iterations (learning rate 7, momentum 0.5), followed by extended training (50,000 iterations, learning rate 0.1, momentum 0.5) to identify the optimal $L_{pixel}$ within the bounds [-1,0] and [0,1].
For ImageNet \cite{deng2009imagenet}, we tested FoCLIP on 25–100 randomly selected classes (sampling per \cite{FreibergeCLIPMasterPrints}) using ViT-L/14, with identical parameters (1,000 iterations, learning rate 7, momentum 0.5).\\
\indent To evaluate robustness to grayscale conversion, we compared score degradation between the original and our fooling examples. For generalization testing, we selected 25 ImageNet classes\cite{FreibergeCLIPMasterPrints}, used 25×50 images for generalization verification, trained ViT-L/14 for 1,000 iterations, converted the results to grayscale and visualized score changes via density maps. Finally, we validated our double-threshold detection mechanism on these 25 classes.

\subsection{Performance of FoCLIP}

\begin{figure}[htbp]
  \centering
  \includegraphics[trim=205 65 190 45, clip,width=1\textwidth, page=2]{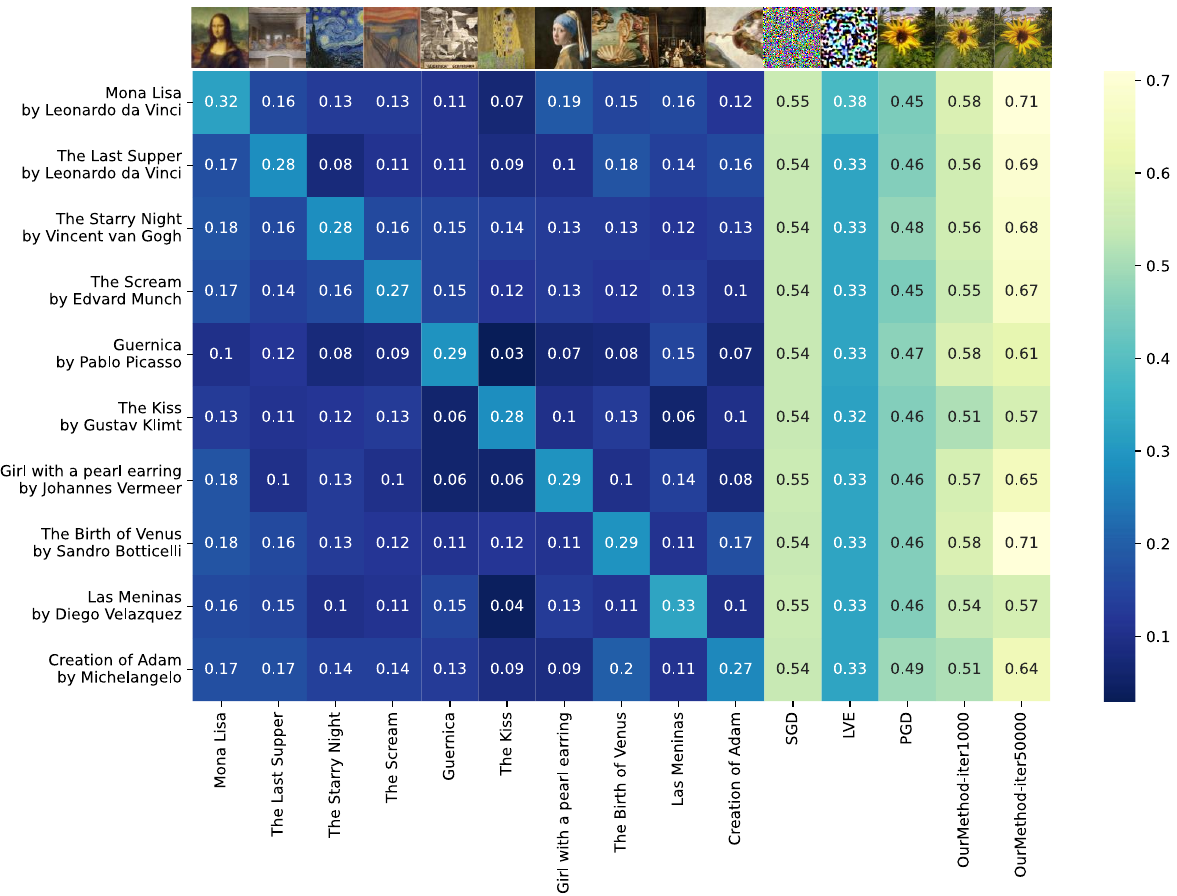} 
  \vspace{-0.8cm}
  \caption{Heatmap of CLIPscore of famous artworks and titles, including CLIPMasterPrints for SGD, LVE and PGD approaches\cite{FreibergeCLIPMasterPrints}, and comparing with our methods with 1000 and 50,000 iterations. Our fooling examples showed the best performance.}
  \label{fig:heatmap}
  \vspace{-0.5cm}
\end{figure}

\begin{figure}[htbp]
  \centering
  \vspace{-2.6cm}
     \begin{minipage}[b]{.45\linewidth} 
     \centering
     \begin{picture}(200,200)
       \put(0,120){(a)}
       \put(0,0){\includegraphics[trim=220 100 220 50, clip,width=1\textwidth, page=3]{pictures.pdf}} 
     \end{picture}  
     \end{minipage}
    \begin{minipage}[b]{.45\linewidth}   
    \centering
    \begin{picture}(200,200)
       \put(0,120){(b)}
       \put(0,0){\includegraphics[trim=0 -50 0 0, clip,width=.9\textwidth, page=1]{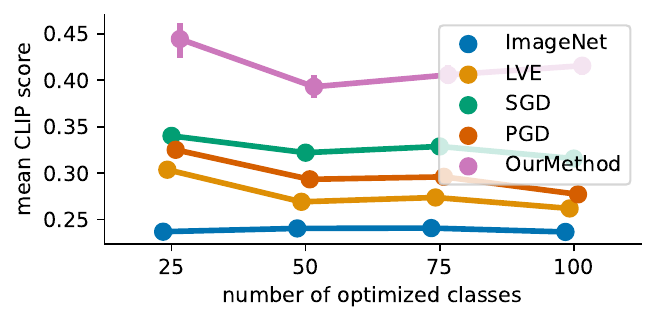}} 
    \end{picture}
    \end{minipage}  
  \caption{(a) CLIPscore comparison of fooling images generated by SGD, LVE, PGD and our method across 25 target classes, alongside similarity scores of corresponding ImageNet validation images. (b) Average similarity trends across 25-100 categories show our method outperforms others significantly, with minimal score degradation as category count increases (note: some variance values are imperceptible due to scale in (b)).}
  \label{fig:Scatter plots and Line charts}
  \vspace{-0.5cm}
\end{figure}

\noindent As shown in Fig.\ref{fig:heatmap}, the results of the SGD, LVE and PGD methods in the figure are reproduced based on Matthias Freiberger et al.\cite{FreibergeCLIPMasterPrints}. It can be proved that our method shows a stronger attack ability when facing the modality gap between text and image embedding, and at the same time it can ensure a better visual quality of the image. Meanwhile, we verified that when $iter = 1000$, $lr = 7$ is the best choice. And it can be seen in Fig.\ref{fig:Scatter plots and Line charts} that when both methods are run for 1,000 iterations, our approach consistently outperforms the best method (SGD) of the original paper. In particular, while SGD generates noise-like artifacts, our method demonstrates a significant advantage in terms of deceptive capability.\\

\begin{figure}[H]           
\centering    
\vspace{-0.3cm}
\includegraphics[trim=200 90 200 75, clip,width=1\textwidth, page=1]{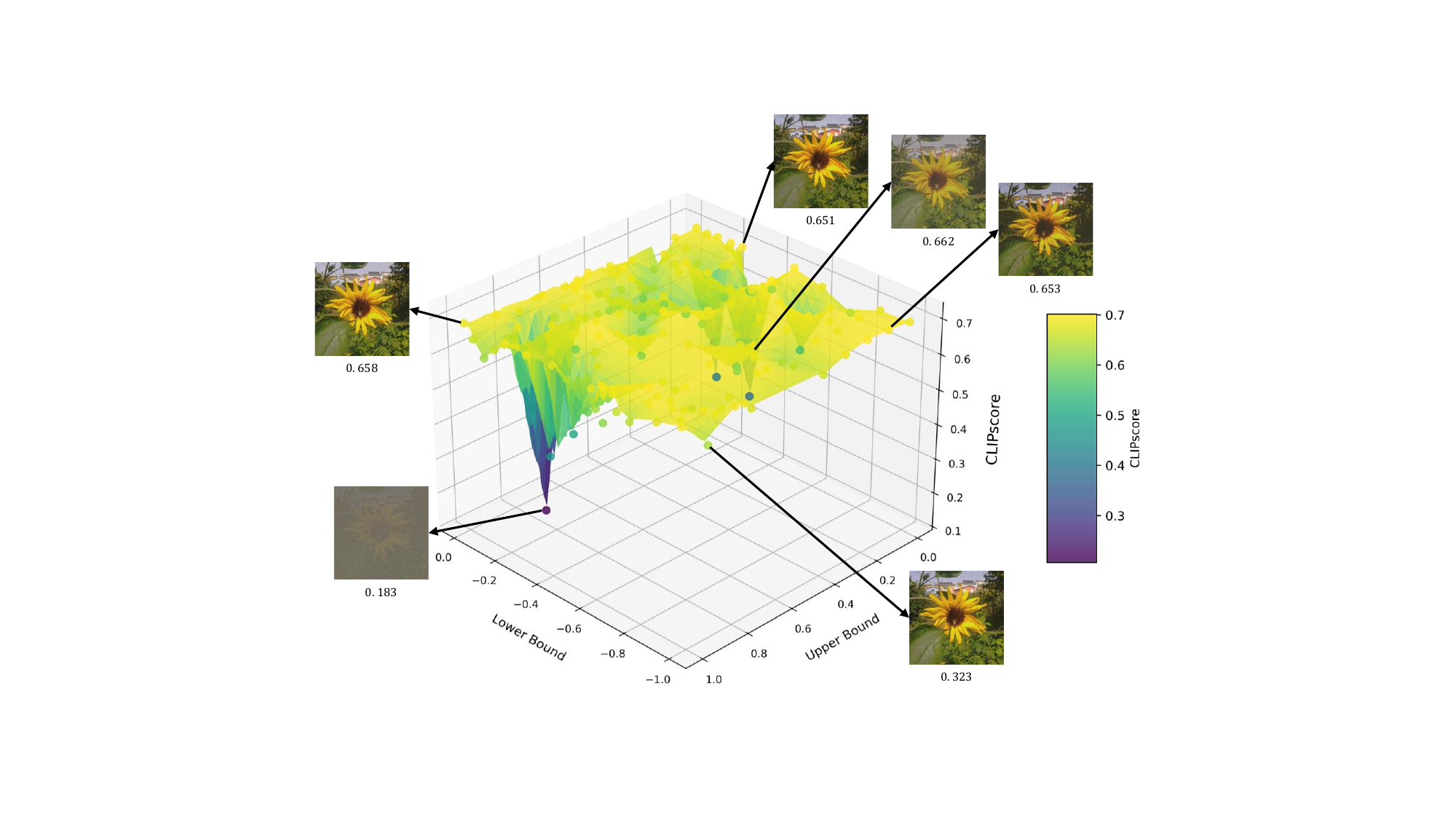} 
\caption{To illustrate the relationship between pixel-guard regularization bounds and CLIPscore, we visualize it via a 3D graph. The x- and y-axes represent $bound_{lower}\in[-1,0]$ and $bound_{upper}\in[0,1]$, while the z-axis indicates CLIPscore. Representative fooling images are displayed at key points.}
\vspace{-0.5cm}
\label{fig:lr-3D}
\end{figure}

\indent After that, we searched for $bound_{lower}$ and $bound_{upper}$ in the parameter space in $\mathcal{L}_{pixel}$ and presented it in the three-dimensional space. As shown in Fig.\ref{fig:lr-3D}, the points near [-0.15, 0.8] with the lowest score, the score is only about 0.2, and the image is very unclear.The point of [-0.06, 0.55], with an average score of 0.32, but it is the clearest image.The image around [-0.17, 0.90] is also very clear, and the average score is about 0.64.The image around [-0.80, 0.50] is also relatively clear, and the average score is about 0.65.The image of [0.00, 1.00] has an average score of nearly 0.66, and it can be seen that it is clearer, not much different from the clearest point, but the score is 30 points higher.In conclusion, it can be considered that using the vicinity of [0, 1] as the bounding is a relatively optimal option.\\

\noindent The experimental results show that not only do the samples close to [0, 1] have relatively high scores (with an average score of nearly 0.66), but also their clarity is similar to that of the original image samples. This indicates that the samples close to [0, 1] achieve a balance between the scoring index and the image quality, our speculation is as follows:\\
\indent 1.The model is likely more sensitive to pixel values within the range [0,1]. Values outside this range may cause an explosion or disappearance of the gradient, while values near the boundaries may help maintain stability during optimization.\\
\indent 2.The feature alignment loss in the scoring function may prioritize pixel value constraints, whereas visual clarity correlates with preservation of high-frequency details. Samples near [0,1] may strike a balance between these conflicting objectives.\\
\indent 3.In digital images, pixel values are typically stored as 8-bit unsigned integers (0-255). In our experiments, samples near [0, 1] may better preserve the original color distribution.

\subsection{Grayscale detection results}

\begin{figure}[htbp]
\vspace{-3cm}
    \begin{minipage}[b]{.45\linewidth} 
    \centering
    \begin{picture}(200,200)
        \put(0,120){(a)}
        \put(0,0){\includegraphics[trim=220 100 220 50, clip,width=1\textwidth, page=6]{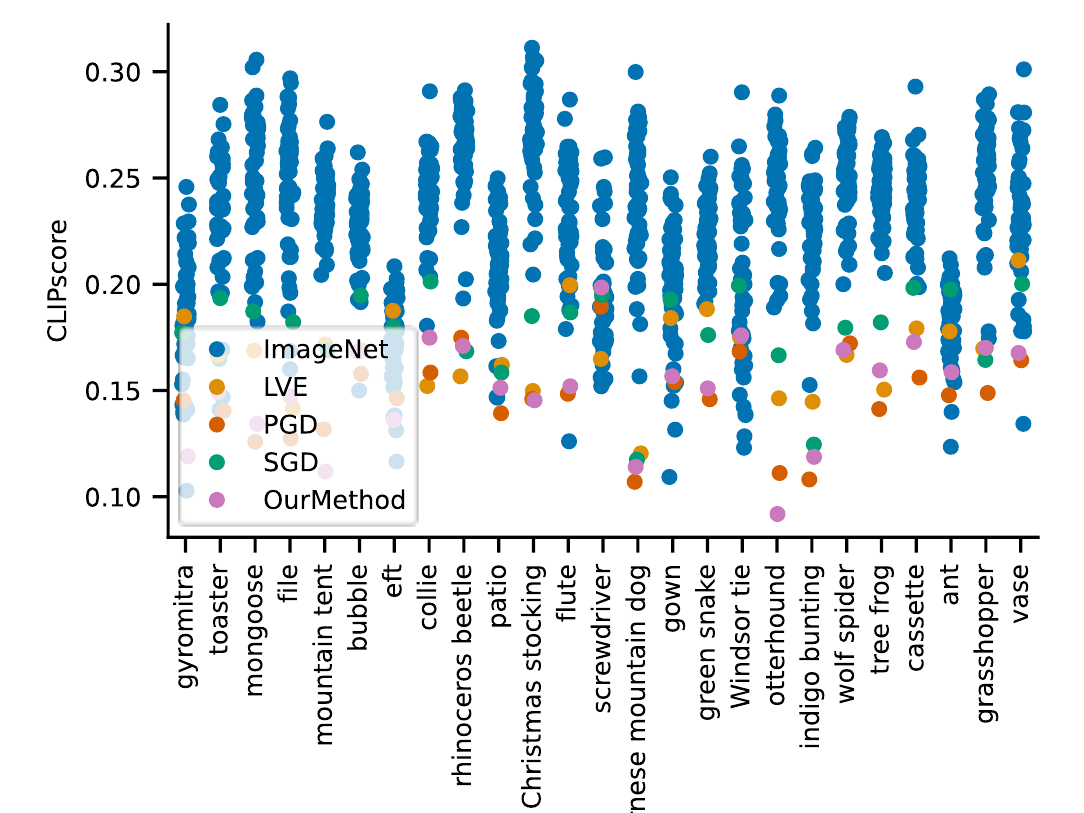}} 
    \end{picture}
    \end{minipage}  
    \begin{minipage}[b]{.45\linewidth} 
    \centering        
    \begin{picture}(200,200)
        \put(0,120){(b)}
        \put(0,0){\includegraphics[trim=250 100 250 100, clip,width=1\textwidth, page=7]{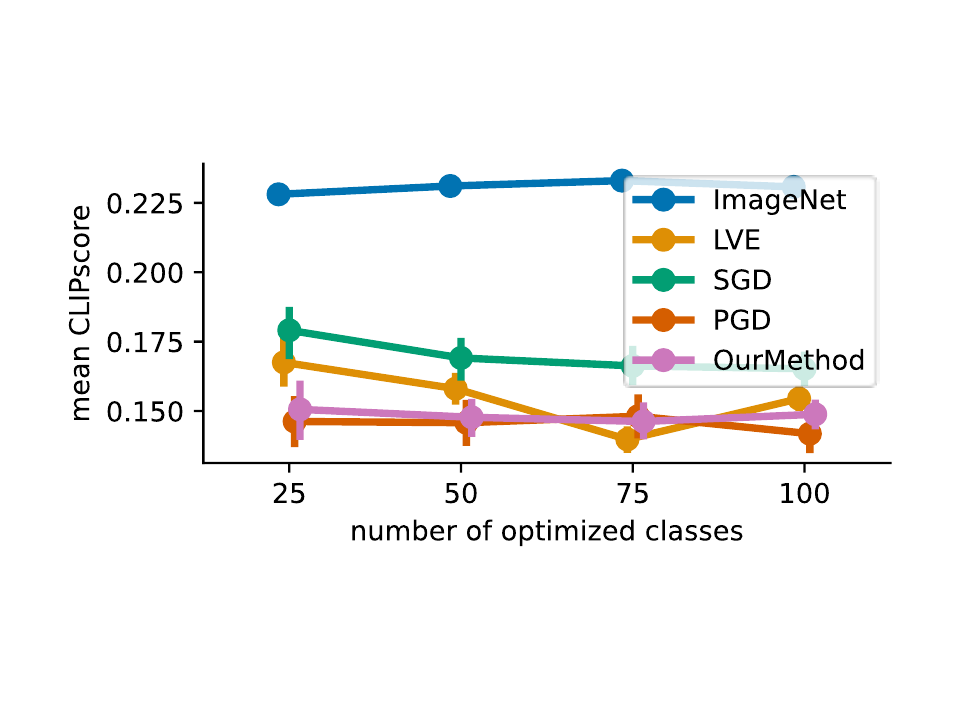}} 
    \end{picture}
    \end{minipage}  
  \caption{ Comparison of four methods and original images using grayscale conversion on images as the same as Fig.\ref{fig:Scatter plots and Line charts}.}
  \label{fig:original verification}
\vspace{-0.5cm}
\end{figure}

As shown in Fig.\ref{fig:original verification}, whether it is the method in the original paper or our method, the CLIPscore of the generated image after grayscale image conversion is significantly reduced. This may be related to the fact that the CLIP model is not sensitive to grayscale images. It is speculated that during pre-training, color information may be an important clue for text alignment, and the model overly relies on color features while ignoring features such as shape and texture that are retained in grayscale.

\begin{figure}[htbp]
\centering    
\vspace{-3.8cm}
\begin{picture}(200,200)
\put(-70,110){(a)}
\put(160,110){(b)}
\put(-75,0){\includegraphics[trim=0 30 0 60, clip,width=1.0\textwidth]{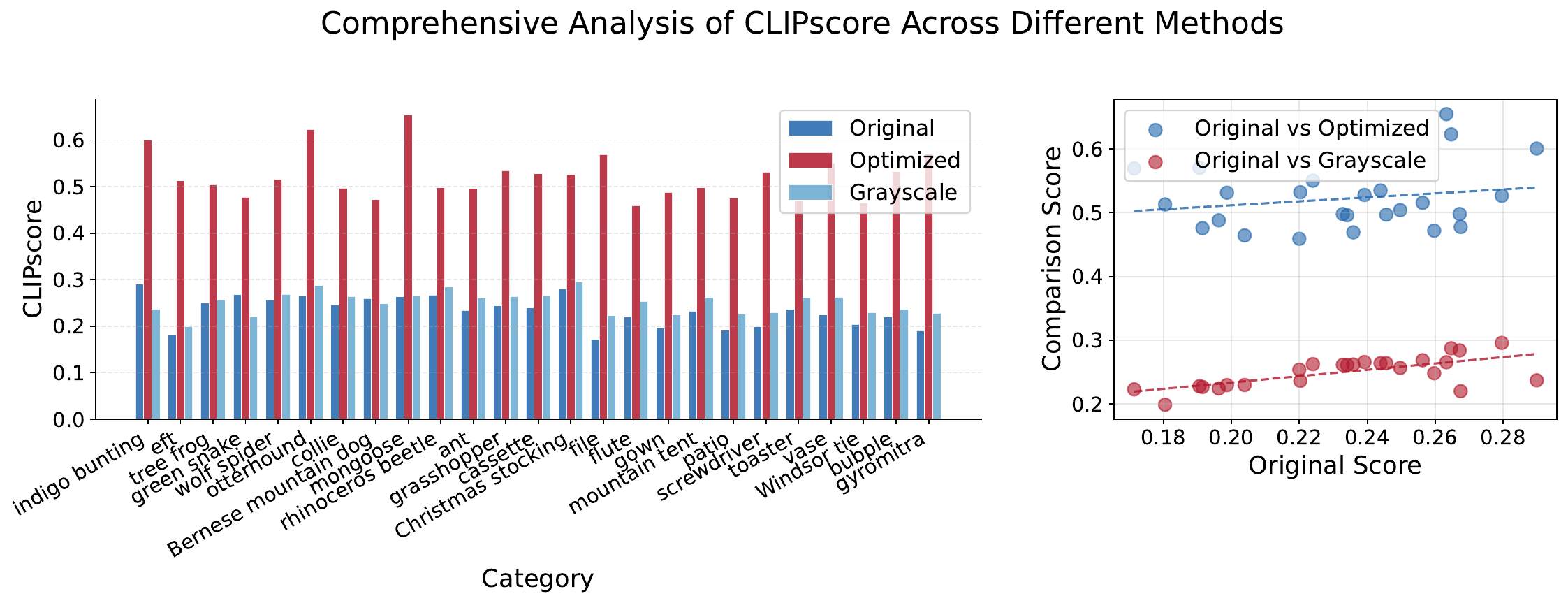}} 
\end{picture}
\vspace{-0.3cm}
\caption{(a) A bar chart showing the average CLIPscore of 50 images in each of the 25 categories. The colors represent the original image, the FoCLIP image with the category as the prompt, and the grayscale image transformed after FoCLIP. (b) The distribution offset after comparing the FoCLIP of the original image and the grayscale image converted after FoCLIP.}  
\label{fig:Generalizability verification}
\end{figure}

\noindent To verify the generalization of FoCLIP and the feature degradation in fooling images, we conducted training and detection on all images of 25 categories, and the training prompt uses the category of the image itself. In Fig.\ref{fig:Generalizability verification}, we verified that no matter how good the modal enhancement effect between the image and the label text is, the CLIPscore after grayscale conversion will significantly decrease, and some scores are even lower than the CLIPscore of the original image. At the same time, the bar chart of the mean CLIPscore more intuitively illustrates the significant distribution difference between the optimized image and the grayscale image score, and also proves the rationality of performing double-threshold detection.\\

\vspace{-0.5cm}
\noindent To verify the effectiveness of the grayscale sensitivity detection mechanism, evaluate its performance in distinguishing between the original image and the optimized and tampered image (fake), analyze the robustness and generalization ability of the detection threshold, we evaluate on the generated images shown in Fig.\ref{fig:Generalizability verification}. The positive sample set is the 25 types of tampered images in the ImageNet verification set, and the negative sample set is the original unoptimized image, which corresponds to the positive sample one by one.

\begin{figure}[H]    
\centering    
\vspace{-3.4cm}
\begin{picture}(200,200)
\put(-75,105){(a)}
\put(45,105){(b)}
\put(170,105){(c)}
\put(-75,0){\includegraphics[trim=0 100 0 160, clip,width=1\textwidth, page=2]{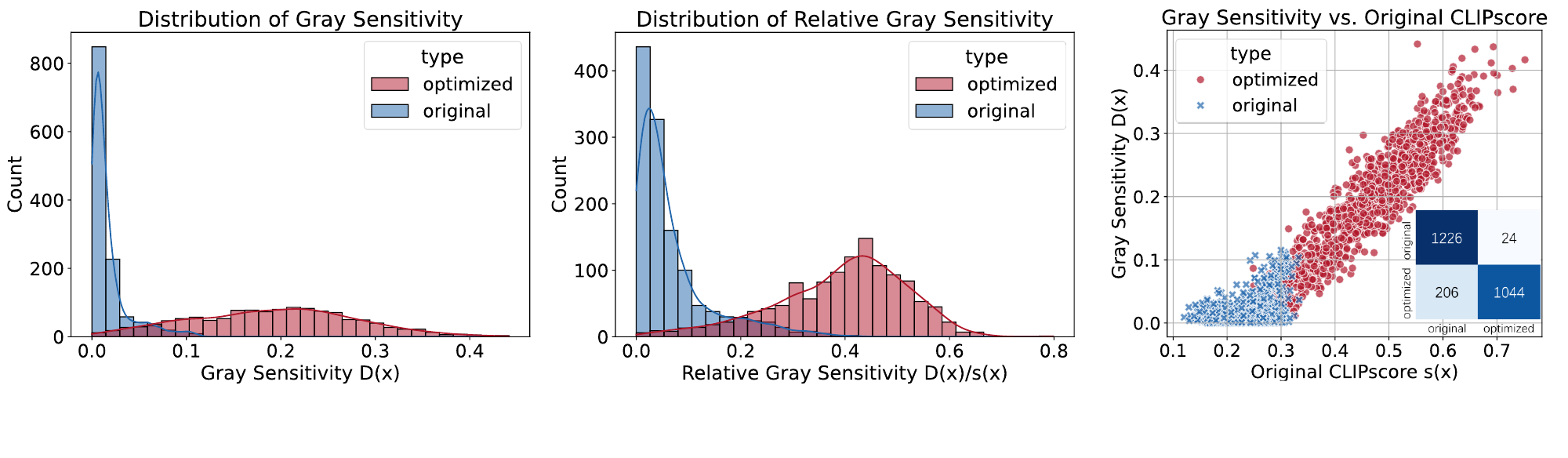}} 
\end{picture}
\vspace{-0.8cm}
\caption{(a) and (b) are the grayscale sensitivity distribution maps of the absolute threshold and the relative threshold. (c) The scatter plot showing the relationship between the original CLIPscore and the grayscale sensitivity. The lower right corner is the confusion matrix using a hybrid method combining absolute and relative thresholds, demonstrating the counts of true positives (TP), false positives (FP), true negatives (TN), and false negatives (FN).}  
\label{fig:grayscale detection}
\end{figure}
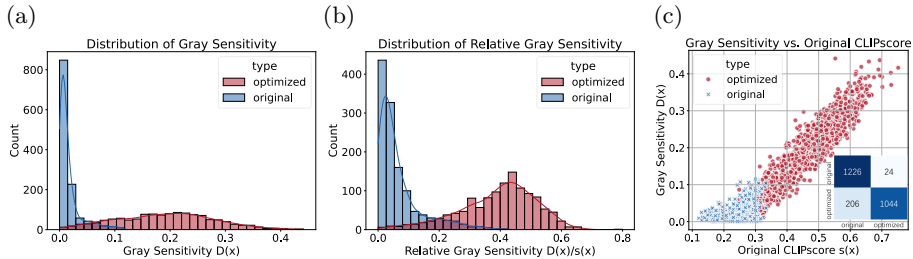

\vspace{-0.5cm}
\noindent As shown in Fig.\ref{fig:grayscale detection}, after analyzing the distributions of $D(x)$, $s(x)$, and $D(x)/s(x)$, iteratively optimizing the values of $\tau_1$ and $\tau_2$, the confusion matrix clearly demonstrates a precision of \textbf{91\%}, highlighting the effectiveness of grayscale sensitivity detection in identifying CLIP-based spoofed images. However, since different adversarial methods exhibit varying CLIPscore distributions, it remains challenging to validate the universal applicability of this approach, particularly when the CLIPscore between spoofed images and original images show minimal differences.

\subsection{Ablation Experiment}
To verify the functions of each part of the FoCLIP method, we conducted an ablation experiment, and the results are as follows:\\
\begin{figure}[H]
  \centering
\vspace{-1.1cm}
  \includegraphics[trim=120 140 0 170, clip,width=1.2\textwidth, page=3]{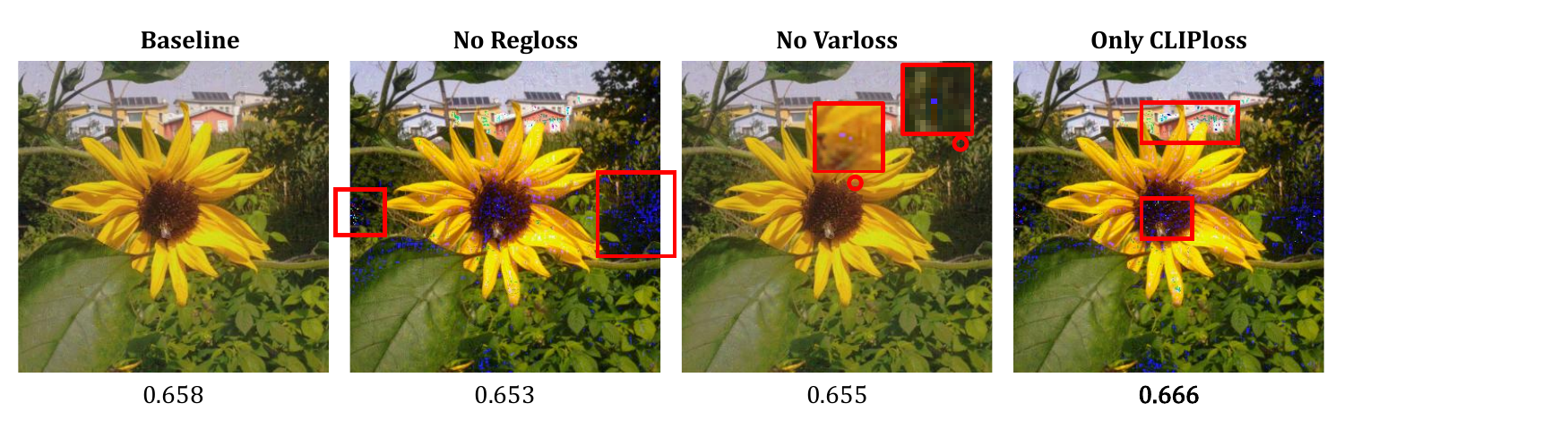} 
  \vspace{-0.3cm}
  \caption{The baseline refers to the FoCLIP method, all scores are average scores. The last three figures show the ablation experiment results of three parts in the FoCLIP method, corresponding to the cases of removing Regloss, removing Varloss, and only retaining CLIPLoss respectively. Since CLIPLoss is the core component of this method, no separate experiment was conducted to remove it.}
  \label{fig: Ablation Study}
  \vspace{-0.5cm}
\end{figure}

\noindent From Fig.\ref{fig: Ablation Study}, it can be concluded that the three parts of this method all have their own functions. While ensuring a relatively high score, it also takes into account the visual quality of the image.

\section{Conclusions}
In this work, we introduced \textbf{FoCLIP}, a multimodal feature misalignment optimization framework aimed at making fooling image to improve CLIPscore and provide a practical tampering detection method within the CLIP model. Our research explored the critical issue of adversarial attacks on pre-trained CLIP model, which have become increasingly sophisticated and pose significant threats to the integrity of digital content. (1) By constructing a multi-objective equilibrium model, improving the CLIP similarity score between the image and the target text (with an average increase of 27.3\%), the visual quality is ensured to be maintained. Experiments show that the average CLIPscore of the optimized image on the artistic masterpiece prompt words has increased by 42.7\%, and the feature similarity of "Mona Lisa" reaches the highest at 0.709. (2) Experiments on the ImageNet dataset show that FoCLIP shows a stable generalization ability for attacks on 25, 50, 75, and 100 types of targets. Especially in multi-category attacks, the fluctuation range of the similarity score is less than ±3.7\%, verifying the universality of the framework. (3) We revealed the vulnerability of the CLIP model to grayscale conversion: the CLIPscore of the optimized image decreases by an average of 63.2\% after grayscaling, while the score of the original image only decreases by 8.5\% Based on this phenomenon, the designed double-threshold detection mechanism achieves the highest detection accuracy of 91\% on the standard test set.

%
%
%
%





\end{document}